\documentclass{article}
\usepackage{ijcai20}

\usepackage{times}
\usepackage{soul}
\usepackage{url}
\usepackage[hidelinks]{hyperref}
\usepackage[small]{caption}
\usepackage{graphicx}
\usepackage{amsmath}
\usepackage{amsthm}
\usepackage{booktabs}
\usepackage{algorithm}
\usepackage{algorithmic}
\usepackage{multirow}
\usepackage{booktabs}
\usepackage{float}
\title{Weakly-supervised Object Localization for Few-shot Learning and Fine-grained Few-shot Learning}

\author{
Xiaojian He$^1$
\and
Jinfu Lin$^1$\and
Junming Shen$^1$
\affiliations
$^1$South China University of Technology
\emails
\{Jinfu Lin\}kingfoulin@foxmail.com
}

\begin{document}

\maketitle

\begin{abstract}
  Few-shot learning (FSL) aims to learn novel visual categories from very few samples, which is a challenging problem in real-world applications. Many methods of few-shot classification work well on general images to learn global representation. However, they can not deal with fine-grained categories well at the same time due to a lack of subtle and local information. We argue that localization is an efficient approach because it directly provides the discriminative regions, which is critical for both general classification and fine-grained classification in a low data regime. In this paper, we propose a Self-Attention Based Complementary Module (SAC Module) to fulfill the weakly-supervised object localization, and more importantly produce the activated masks for selecting discriminative deep descriptors for few-shot classification. Based on each selected deep descriptor, Semantic Alignment Module (SAM) calculates the semantic alignment distance between the query and support images to boost classification performance. Extensive experiments show our method outperforms the state-of-the-art methods on benchmark datasets under various settings, especially on the fine-grained few-shot tasks. Besides, our method achieves superior performance over previous methods when training the model on miniImageNet and evaluating it on the different datasets, demonstrating its superior generalization capacity. Extra visualization shows the proposed method can localize the key objects more interval.
\end{abstract}

\section{Introduction}

Deep Convolutional Neural Networks (ConvNets) has achieved excellent performance in numerous computer vision tasks in recent years. Trained with a large amount of annotated data, the ConvNets can extract robust and effective representations for classification. However, ConvNets suffers from its weak generalization ability and poor performance when the annotated samples for training are very limited. In contrast, we humans can identify novel classes with only a single or few samples. Thus, recognizing novel categories from very few samples is an important and significant problem, which is often termed Few-shot learning (FSL).

\begin{figure*}
  \centering
  \includegraphics[width=0.75\textwidth]{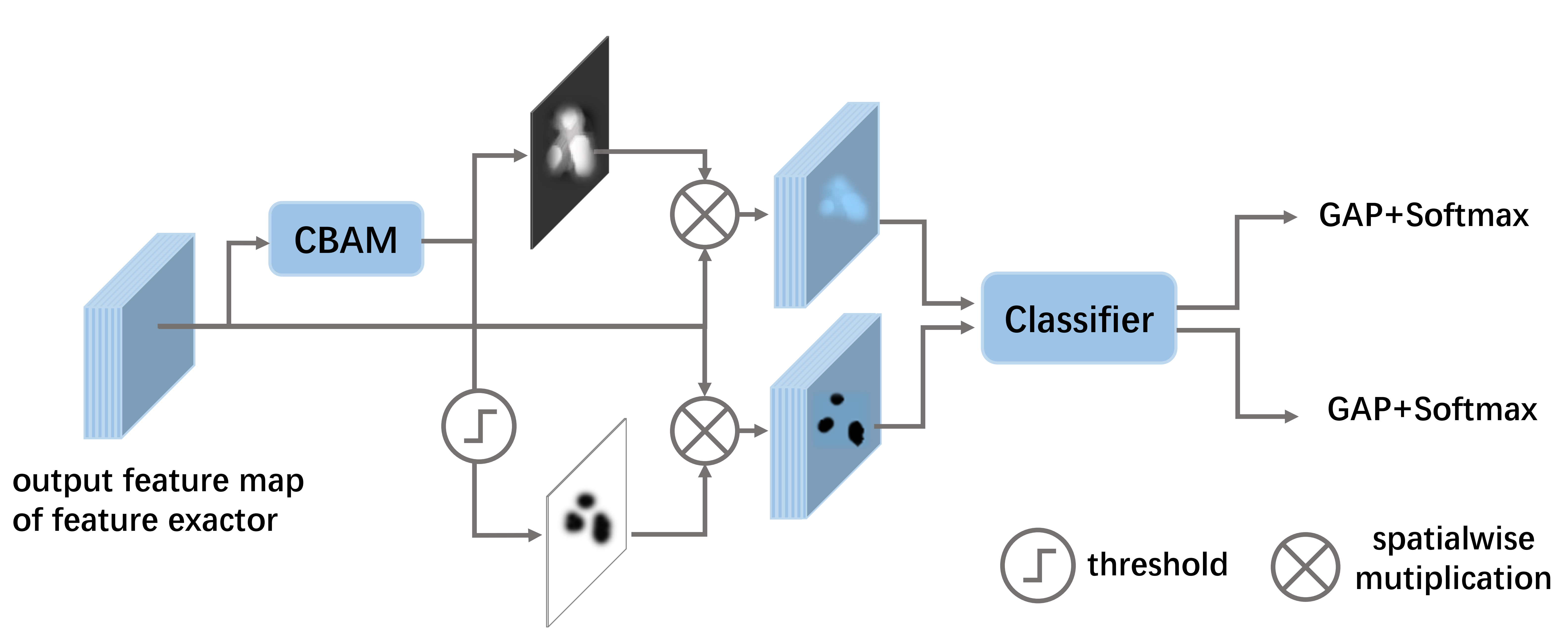}
  \caption{Illustration of Self-Attention Based Complementary Module (SCA module). Our method mainly contains: (a) Channel-Based Attention Module (CBAM) to generate the important mask, (b) Classifier to fulfill the classification, and output the class activation map (cam). The classifier learns to obtain two complementary discriminative regions by classifying two complementary features to the same classes. The complementary features contain the features spatial-wise multiplication with the important mask and the features spatial-wise multiplication with the complementary erased mask. Note that our SAC module is lightweight can be easily applied to convolutional feature maps of various models (e.g., VGG16) to improve the localization accuracy. Best viewed in color.}
  \label{}
\end{figure*}

Recently, the FSL problem has attracted increasing attention, and a number of methods have been proposed to tackle this task. Fine-tuning the pre-trained model on the novel datasets is a common and simple method. Besides, To generalize the model to novel datasets, an emerging direction is to apply the meta-learning paradigm on few-shot learning. Meta-learning trains an across-task meta-learner which can accumulate transferable knowledge in one task and generalize to other novel tasks quickly. Another common approach is based on metric-learning, which learns an informative similarity metric between the query and the support samples, thus performing few-shot classification.

Due to the very limited samples, most of the approaches encounter over-fitting and poor generalization. Thus, many data augmentation (DA) methods have been proposed. \cite{Wang2018low} and \cite{Hariharan2017low} are both data generation based method, which can generate additional examples for data-starved classes. However, they need a large of extra annotated data to train such a specialized data generator or hallucinator.  \cite{Schwartz2019} leverages extra multiple semantic and feature fusion to train a more robust embedded module. Nevertheless, these methods contain complicated feature fusion networks, and refer to extra semantic information may limit application scenarios.

Different from DA, localization can distinguish the most discriminative regions from distractors without using extra annotated samples.  Mask-CNN \cite{wei2016mask} utilized the fully convolutional network to locate the most discriminative parts to fulfill the fine-grained recognition. Inspirited by this, we argue that guiding to localize the discriminate regions to perform few-shot classification should make a significant improvement of the FSL task, especially the fine-grained few-shot (FGFS) task. However, many weakly-supervised object localization (WSOL) methods fail to localize the integral regions of the objects. For instance, \cite{Zhou2016,Oquab2015} replace the last few layers of the classification network with a global pooling layer and a fully-connected layer to generate the discriminative class activation maps (CAM). However, CAM tends to cover only the most discriminative part which leads to classification accuracy improvement.

To bridge this gap, we propose a novel end-to-end network to achieve weakly-supervised object localization which is shown in Figure 1. Inder to localize the integral objects and select the most discriminate features for few-shot classification, we design the Channel-Based Attention Module (CBAM). CBAM takes the feature maps as input and generates the important mask. The complementary erased mask is also produced using the threshold. Then the classifier classifies the complementary features to the same class to obtain the completely class activation map (e.g., $CAM_{imp}, CAM_{erased}$). By fusing both the $CAM_{imp}$ and $CAM_{erased}$, the SCA module can capture the more interval class activation map for the input images. After obtaining the most discriminative regions, we applied the interpolation to select the useful deep descriptors and feed them to the semantic alignment module to compute the semantic alignment distance for few-shot learning.

Many previous methods compute a prototype of each class by averaging all the support data of the class. Despite its efficiency, it is vulnerable to noise. Inspirited by NBNN \cite{Boiman2008}, we align each deep descriptor of the key object by its nearest neighbor among all of the support deep descriptors, which mean the distance between the query images and the support images can be computed in a relative high-data regime, making it more stable and biased-noisy.
We encapsulate this part as a semantic alignment module to output the distance. The pipeline of our method for few-shot classification is shown in Figure 2.

Below, we list our main contributions: (1) We propose a lightweight and efficient module named SAC module that can localize the integral discriminative region. The designed CBAM efficiently helps the classifier to capture the discriminative part and complementary discriminative part. (2) We design the semantic alignment module to boost few-shot classification over the selected deep descriptors since it effectively reduces background noise. (3) Extensive experiments on benchmark datasets and fine-grained few-shot datasets, and the generalization evaluation experiment all show the superiorities of our method. Meanwhile, the visualization shows the proposed method can localize the key objects accurately.

\section{Related Work}
\subsection{Meta-learning and Metric-learning}
Meta-learning based method trains an across-task meta-learner with the meta-learning paradigm. MAML \cite{Finn2017} trained a model agnostic meta-learner and found the initial parameters adapting to a variety of tasks with similar distribution, such that the model can quickly generalize to the new tasks. Meta-Learning LSTM \cite{Ravi2016} proposed a model based on LSTM to learn an optimization method as well as the general initialization of the classifier. Metric-learning tackles the FSL by learning an embedding space where the input of the samples of the same categories is closer than those of different categories. Combining with attention mechanism, Matching Network \cite{Vinyals2016} used the cosine distance to train a k-nearest neighbor classifier on the learned embedding space. \cite{Snell2017} proposed a prototypical network to learn a prototype representation of each category and performed classification by the Euclidean distance between the query and prototype. Relation network \cite{Zhang2018} learned a nonlinear comparator to compare the distance metric between the query and the support images. Different from these metric-learning approaches that directly using the vector obtained by flattening the embedded feature, we use a set of selected deep descriptors to represent the embedded feature.

\subsection{Object Localization}
Many works \cite{Wei2017,Wei2017} have shown that utilizing localization can help to learn the more discriminative embedded feature for classification. In this context, \cite{Wei2017} presented the Selective Convolutional Descriptor Aggregation method to achieve unsupervised localization in fine-grained datasets. Similarly, \cite{Sun2019} used the class attention map (CAM) generated by classification networks to locate the key region and fused with other different scale features for fine-grained few-shot classification.  \cite{Wertheimer2019} used the assistant bounding box annotations to achieve the localization within the few-shot classification, thus to address the FSL task over heavy-tailed datasets. Leveraging localization can improve few-shot classification performance. However, in real-world applications, bounding box annotations may be hard to meet or impracticable. To address the FSL problem with realistic settings, we propose a novel method to achieve weakly-supervised object localization.

\section{The Proposed Model}

\subsection{Problem Definition}

There is the train dataset $D$, support dataset $S$, and the query dataset $Q$ in the FSL task. $D=\left\{\left(x_{i}, y_{i}\right)\right\}_{i=1}^{N}$ contains $N$ samples, where $y_i$ is the label of image $x_i$.  The support set $S=\left\{\left(x_{j}, y_{j}\right)\right\}_{j=1}^{M}(M=C * K)$ includes $M$ examples in test phase and there is $K$ labeled samples for each of $C$ novel categories (C-way K-shot problem). The query dataset $Q=\left\{\left(x_{j}, y_{j}\right)\right\}_{j}^{N_{q}}$  shares the same label space with $S$. Their relationship is denoted as  $(S \cup Q) \cap D=\emptyset$. FSL aims to train a model from $D$, then classify the novel samples from $Q$ based on the $S$ during the testing phase. To mimic the few-shot learning task, the episodic training mechanism is adopted to train the model. Episode is a mini-batch includes $D_{support}$ and $D_{query}$, where we randomly sample $C$ categories from $D_{train}$ and for each category of $C$ categories,  its labeled samples are randomly split into subset $D_{support}$ with $K$ samples and subset $D_{query}$ with the rest samples. Through this training mechanism, the model can learn transferable knowledge.

\subsection{Model}
\begin{figure*}
  \centering
  \includegraphics[width=0.85\textwidth]{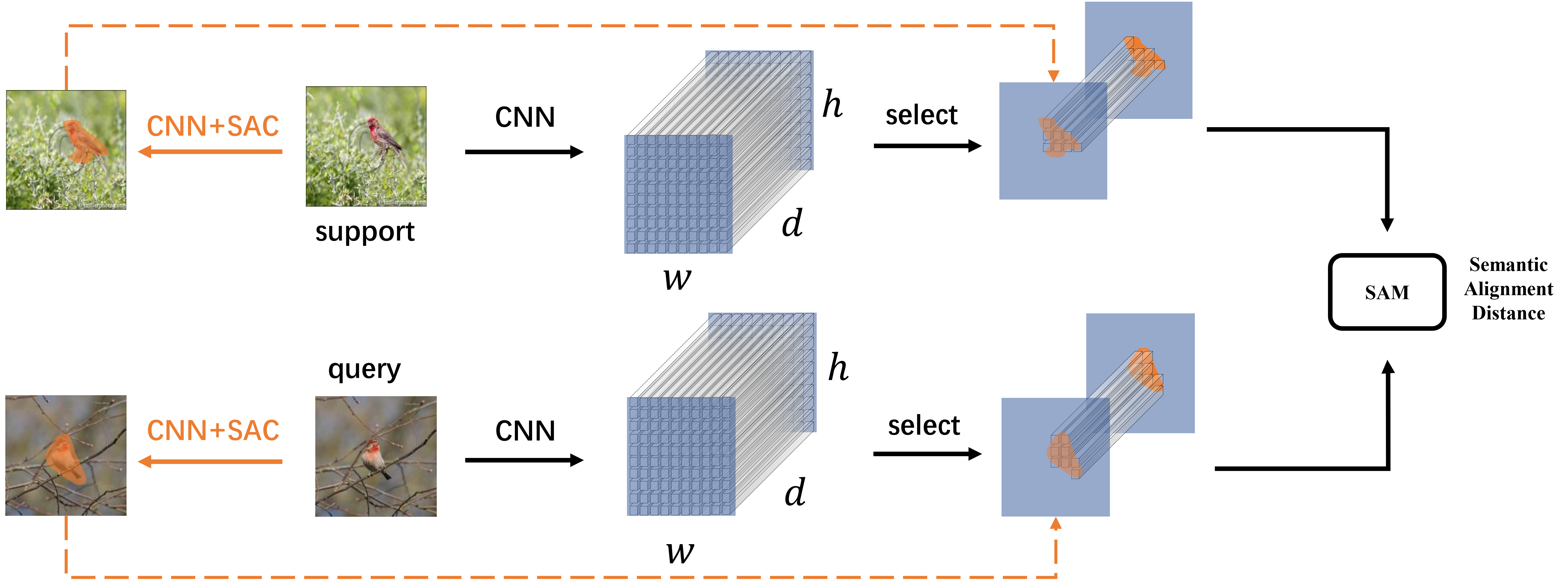}
  \caption{The architecture of our method for few-shot learning.  We utilize Conv-64 or ResNet-12 as the backbone to obtain the convolutional activation tensor. Then we use the fused cam generated by the WSOL network (VGG16+SAC) to select the useful deep descriptors. The cam is resized to a suitable scale by nearest interpolation. Excepted the selected deep descriptors, the rest deep descriptors are set to zero vector. Finally, the selected is feed to the SAM module to compute the semantic alignment distance.  We maximizing the distance if the input pair belongs to the same class while minimizing the score if comes from a different class. Best viewed in color.}
  \label{}
\end{figure*}

\paragraph{Deep descriptor}
For an image $X,$ the activation of a convolution layer can be formatted as an 3D tensor denoted as $E(X) \in R^{d \times w \times h} .$ On the one hand, $E(X)$ includes $d$ feature maps with the size of $w \times h$ and is denote as $M=\left\{M_{n}\right\}(n=1,2,3, \cdots, d), M_{n}$ also known as the feature map in $n t h$ channel. On the other hand, $E(X)$ can be considered as including $m=(w \times h)$ deep descriptors and each deep descriptor is a $d$ -dimension vector. We denote it as:
\begin{align}
  \resizebox{.91\linewidth}{!}{$
  D=\left\{d_{(1,1)}, d_{(1,2)}, \cdots, d_{(i, j)}, \cdots, d_{(h, w)}\right\}=\left\{d_{1}, d_{2}, \cdots, d_{m}\right\}
  $}
\end{align}
where $(i, j)$ is the position of the descriptor and $d_{(i, j)} \in R^{d}$. Thus, a set of deep descriptors is the representation containing spatial information.

\paragraph{Channel-Based Attention Module (CBAM)}
From previous works, erasing the most discriminate part is the effective method to obtain the integral object in WSOL tasks. We design a very lightweight yet efficient module to capture the most discriminative part based on the channel attention, which is denoted as CBAM. Through the CBAM, we produce an important mask directly. Then a complementary erased mask is also produced by the threshold. The module can be easily incorporated into various existing models (e.g., vgg16).

\begin{figure}[H]
  \includegraphics[width=0.48\textwidth]{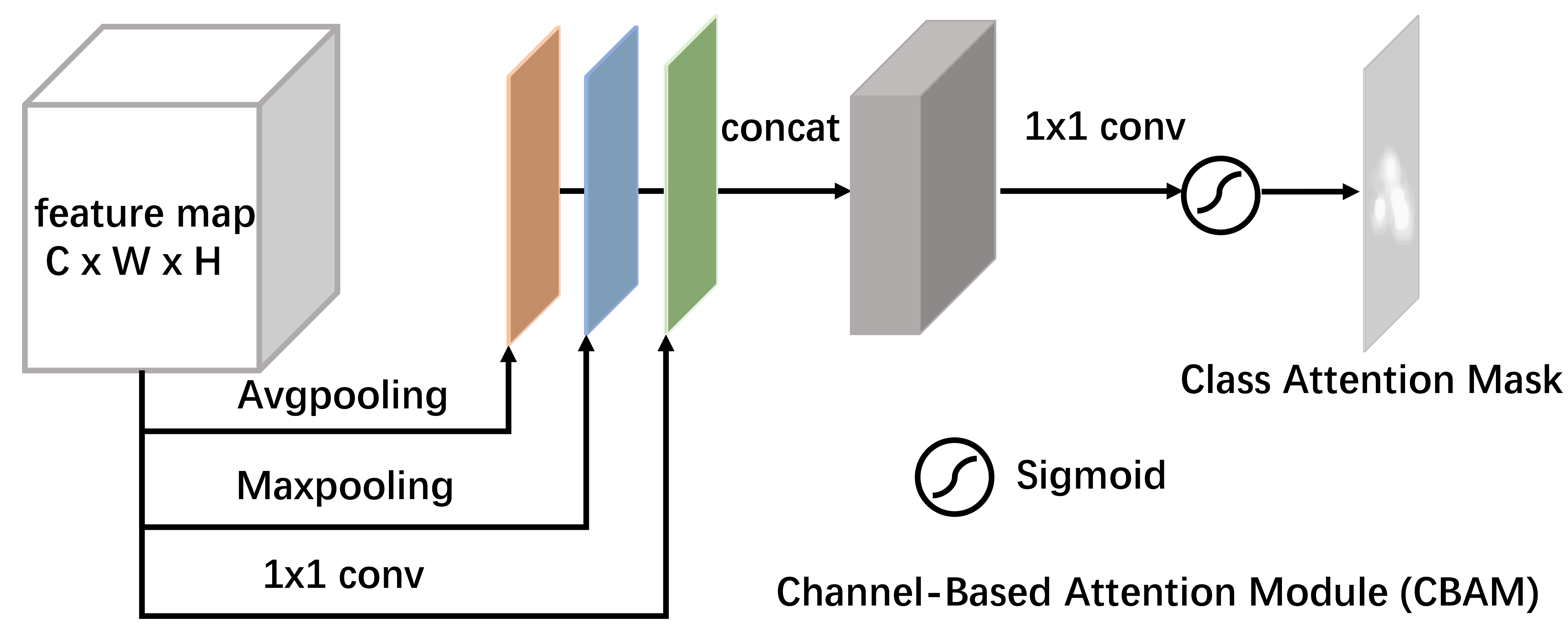}
  \caption{Illustration of the CBAM. It contains the avgpooling, maxpooling, and 1x1 convolution operation. Then we concat the output of the three operations and apply a 1x1 convolution operation with a sigmoid function to produce the important mask of the input feature maps. Best viewed in color.}
  \label{}
\end{figure}

\paragraph{Classifier to obtain the cam}
Different from CAM which needs an extra step (e.g. CAM get the classification weight through gradient backhaul) to obtain the class activation maps after the forward, Acol \cite{zhang2018adversarial} proposed a novel method to obtain the class activation map directly from the feature map of the last convolutional layer. Suppose there are C classes during the meta-train phase, the last convolutional layer of the classifier is a C channel with 1x1 kernel size. The output of the classifier is fed to the softmax for classification. Suppose the weight matrix of the 1x1 convolutional layer is $W^{1 \times 1}\in R^{K \times C}$. Then we can directly obtain the class activation map:

\begin{equation}
  \begin{aligned}
    A_{c}^{cam} = \sum_{k=0}^{k=K-1}S_k \dots W_{k,c}^{1 \times 1}
  \end{aligned}
\end{equation}

To obtain the integral object localization map, we fuse the two complementary ($CAM_{imp}$, $CAM_{earsed}$) by max operaton:
\begin{equation}
  CAM = max(CAM_{imp},CAM_{erased})
\end{equation}

\begin{figure}[H]
  \includegraphics[width=0.45\textwidth]{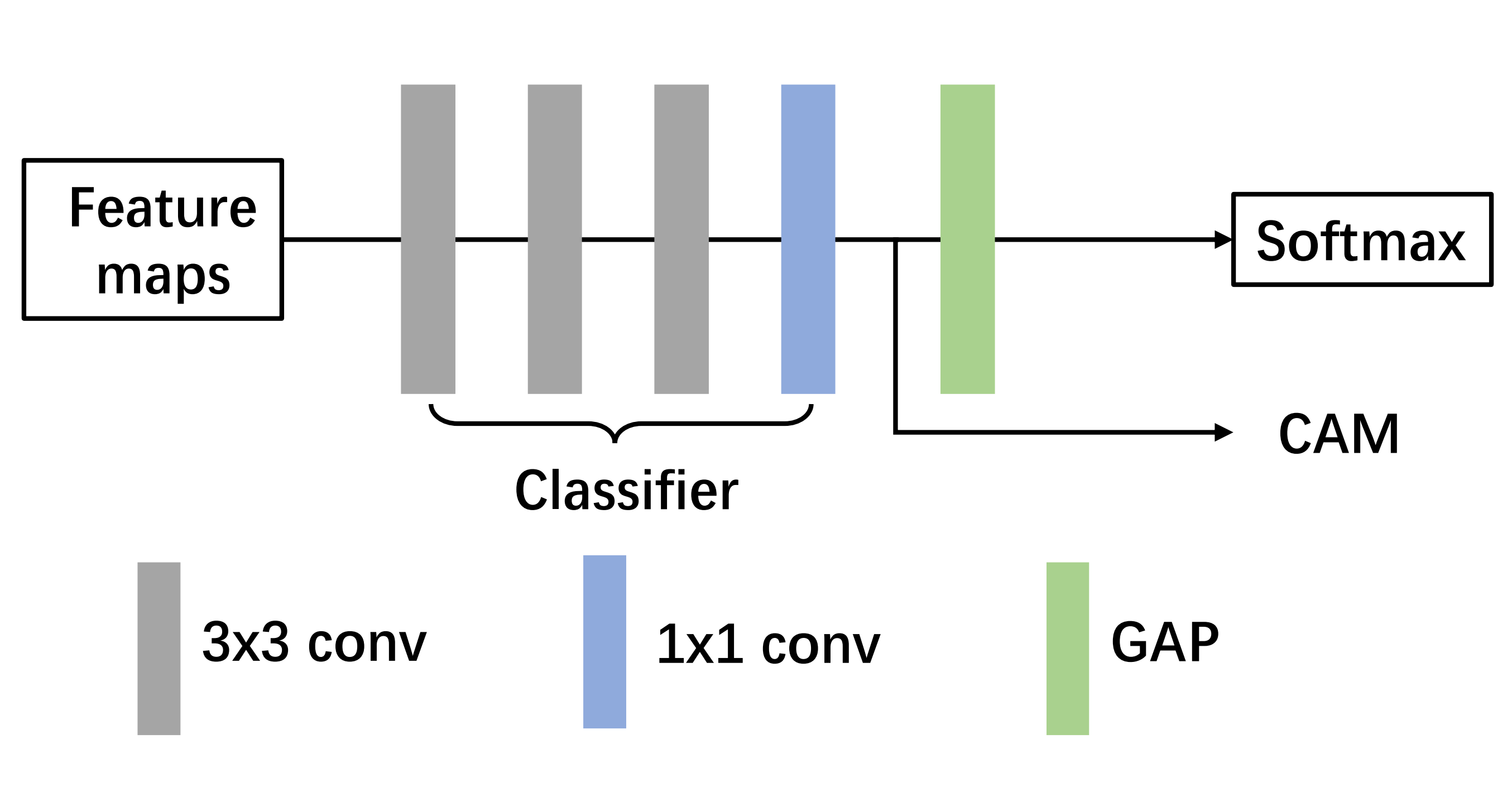}
  \caption{Illustration of the process to produce the cam through the classifier. The classifier contains 3 $3 \times 3$ convolutional block  and 1 $1 \times 1$ convolutional block. Best vied in color. }
  \label{}
\end{figure}

\paragraph{Semantic Alignment Module (SAM)}
This module is designed to calculate the semantic alignment distance for the query/support pair. In this paper, we suppose that each deep descriptor of the key object is independent and has clear semantics. For instance, the selected set of deep descriptors can be interpreted as a bird, dog, etc. Based on the above suppose, we search the nearest-neighbor (NN) among the support set for each descriptor of the query image and accumulate them as the final distance. We define such distance between two discriminative regions as semantic alignment distance. By applying the NN algorithm over each deep descriptor can we guarantee the content between the query image and support images to be aligned. We chose the cosine distance to measure the distance between two descriptors $d_{i}$ and $d_{j}:$
\begin{align}
  \cos \left(d_{i}, d_{j}\right)=\frac{d_{i}^{T} d_{j}}{\left\|d_{i}\right\|\left\|d_{j}\right\|} \in[-1,1], \quad\left(d_{i}, d_{j} \in R^{d}\right)
\end{align}

For the query image of class $k$, its embedded features is denoted as: $q_{k}=\left\{d_{1}, d_{2}, \cdots, d_{m}\right\} \in R^{d \times m}$. Similarly, the deep descriptors of all support
embedded features of class $k$ are denoted as: $s_{k}=\left\{d_{1}^{\prime}, d_{2}^{\prime}, \cdots d_{l=K \times m}^{\prime}\right\} \in R^{d \times l}$. In order to correctly classify the query image, the model needs to guarantee the semantic alignment distance between $q_{k}$ and its support embedded feature set $s_{k}$ to be highest (nearest), thus that each deep descriptor of the query image can be accurately aligned by its nearest neighbor deep descriptor ($NN\left(d_{i}\right)$) from support set. The semantic alignment distance between the embedded feature of the query image and the embedded feature of the support category $k$ is :
\begin{equation}
\resizebox{.91\linewidth}{!}{$
    \displaystyle
  D\left(q_{k}, s_{k}\right)=\sum_{i=1}^{n}\left\|d_{i}-NN\left(d_{i}\right)\right\|=\sum_{i=1}^{n} NN_{-} \cos \left(d_{i}, \widehat{d}_{i}^{\prime}\right)
  $}
\end{equation}%
where
$\quad NN_{-} \cos \left(d_{i}, \widehat{d}_{i}^{\prime}\right) \quad$ represents $\quad \widehat{d_{i}^{\prime}} \quad$ is the nearest neighbor descriptor of $\quad d_{i}$ among $\left\{d_{1}^{\prime}, d_{2}^{\prime}, \cdots d_{l=K \times m}^{\prime}\right\}$ over cosine distance.
Since the deep descriptors from the key regions of the query will be mostly activated by the deep descriptors from the regions belonging to the same class, Maximizing the score can guarantee each deep descriptor can be aligned correctly. We directly perform classification by the class of its nearest $D$ distance without additional supervisory loss.  For the $C$ - Way $K$ -Shot task, the probability of query image $(x, y)$ belongs to the true category $ k \in \{0,1,2, \cdots, C-1\}$ is denoted as:
\begin{align}
  p_{k}=p(y=k | x)=\frac{\exp \left(D\left(q_{k}, s_{k}\right)\right)}{\sum_{k^{\prime} \in C} \exp \left(D\left(q_{k}, s_{k^{\prime}}\right)\right)}
\end{align}
For $N$ query images $\left\{\left(x_{1}, y_{1}\right),\left(x_{2}, y_{2}\right), \cdots,\left(x_{i}, y_{i}\right), \cdots, \right. \\   \left. \left(x_{N}, y_{N}\right)\right\}$ in each episode, we minimize the loss
$L_{e}:$
\begin{align}
  L_{e}=\sum_{i=1}^{N}-\log p\left(y=y_{i} | x_{i}\right)
\end{align}

\section{Experiments}

\subsection{dataset}
\paragraph{miniImageNet}
 As the mini-version of ImageNet, it contains 100 classes with 600 color images per class. The splits proposed in Matching Net is becoming the standard splitting rule of miniImageNet, where 64 categories are for training,16 categories for validation, and 20 categories for testing. In this work, we adopt this split to compare our approach with state-of-the-art methods.
\paragraph{Fine-Grained Datasets}
In this paper, we pick three fine-grained datasets, e.g., Stanford Dogs \cite{Khosla2011}, Stanford Cars \cite{Makadia2014} and CUB-200 \cite{Wah2011} to conduct the fine-grained few-shot learning task. Stanford Dogs contains 120 categories with 20,580 color images, where 70, 20, and 30 categories are used for training, validation, and testing, respectively. Stanford Cars containing 16,185 color images of 196 classes of cars, which is divided into 130, 17, and 49 categories for training, validation, and testing, respectively. CUB-200 is an image dataset with 6033 color images of 200 bird species for the FGVC task. Similarly, we split it into 130, 20 and 50 categories for training, validation, and testing, respectively.

\subsection{Settings and Experiments}

\paragraph{Settings}
We adopt the episodic training mechanism to make the training phase more faithful to the testing phase. For each training episode, besides the $K$ support images in each class, 5-way 1-shot contains 15 query images while 5 -way 5-shot contains 10 query images for each of the $C$ randomly sampled categories. To be specific, for the 5-way 1-shot task, there 5 support images and 15 query images per class, thus that each episode contains $5 \times 1=5$ support images and $15 \times 5=75$ query images totally. Similarly, for the 5-way 5-shot task, there are $5 \times 5=25$ support images and $10 \times 5=50$ query images totally. In addition, we resize all the input images to $84 \times 84 .$ During the training phase, we randomly sample $300,000$ episodes and select Adam as the optimizer with an initial learning ratio $5 \times 10^{-2}$ which will be reduced by half for every $100,000$ episodes to train our model. During the testing phase, we also randomly sample 600 episodes from the test set to evaluate our model. We adopt the mean accuracy with 95\% corresponding confidence interval as the performance indicator. It is worth mentioning that all our model is trained from scratch in an end-to-end manner, without any finetuning in the test phase.

\paragraph{Few-shot Classification on miniImageNet}
We report the experiment results in table 1. When adopting the Resnet as the embedding module, our model can achieve state-of-the-art results both in the 5-way 1-shot and 5-shot task, especially in the 5-shot task (3.29 \% higher than the 74.44\% reported by DN4 \cite{Li2019a}). Besides, when using the Conv as embedding module, our model also achieves the highest accuracy on the 5-way 5-shot task, gaining the 4.40\%, 1.03\%, and 0.04\% over CovaMNet \cite{Li2019b}, DN4, and SalNet \cite{Zhang2019}. We also obtain very competitive accuracy on 5-way 1-shot task with $Conv$ embedding module, gaining 3.82\%, 2.13\%, 2.08\% improvement over R2D2 \cite{bertinetto2018metalearning}, CovaMNet, and DN4. As for Dynamic-Net and SalNet on the 5-way 1-shot task, they perform very complicated training steps to obtain state-of-the-art results. The former utilizes a two-stage model and needs to pre-train the model while our approach does not. The latter utilizes the state-of-the-art saliency detection model to generate the saliency map, thus to directly locate the key object. On the contrary, our approach achieves weakly-supervised localization with only image-level. Our approach is more simple but efficient and outperforms over state-of-the-art methods both on 5-way 1-shot and 5-shot.
\begin{table}[h]
  \newcommand{\tabincell}[2]{\begin{tabular}{@{}#1@{}}#2\end{tabular}}
  \centering
  \caption{The mean accuracies of the 5-way 1-shot and 5-shot tasks on the miniImageNet dataset, with 95\% confidence intervals.}
    \scalebox{0.85}{
    \begin{tabular}{lccc}
    \toprule
    \multirow{2}[4]{*}{{\textbf{Model}}}  & \multirow{2}[4]{*}{{\textbf{Embedding}}} & \multicolumn{2}{c}{{\textbf{5-Way Accuracy(\%)}}} \\
\cmidrule{3-4}   \multicolumn{1}{r}{} & \multicolumn{1}{c}{} & {1-shot} & {5-shot} \\
    \midrule

    \multirow{2}[2]{*}{{\textbf{Proto Net}}} & {\textit{Resnet}} & {51.15$\pm$0.85} & {69.02$\pm$0.75} \\
    \multicolumn{1}{r}{}  & {\textit{Conv}} & {49.42$\pm$0.78} & {68.20$\pm$0.66} \\
    \midrule
    \multirow{2}[2]{*}{{\textbf{Relation Net}}} & {\textit{Resnet}} & {52.13$\pm$0.82} & {64.72$\pm$0.72} \\
    \multicolumn{1}{r}{}  & {\textit{Conv}} & {50.44$\pm$0.82} & {65.32$\pm$0.70} \\
    \midrule
    \multirow{2}[2]{*}{{\textbf{R2D2}}} & {\textit{Resnet}} & {51.80$\pm$0.20} & {68.70$\pm$0.20} \\
    \multicolumn{1}{r}{}  & {\textit{Conv}} & {49.50$\pm$0.20} & {65.40$\pm$0.20} \\
    \midrule
    \multirow{2}[2]{*}{{\textbf{DN4}}} & {\textit{Resnet}} & {54.37$\pm$0.36} & {74.44$\pm$0.29} \\
    \multicolumn{1}{r}{} &  {\textit{Conv}} & {51.24$\pm$0.74} & {71.02$\pm$0.64} \\
    \midrule
    \multirow{2}[2]{*}{{\textbf{Dynamic-Net}}}& {\textit{Resnet}} & {55.45$\pm$0.89} & {70.13$\pm$0.68} \\
    \multicolumn{1}{r}{}  & {\textit{Conv}} & {\textbf{56.20$\pm$0.86}} & {\textbf{72.81$\pm$0.62}} \\
    \midrule
    \multirow{2}[2]{*}{{\textbf{Ours}}} & {\textit{Resnet}} & {\textbf{58.11$\pm$0.86}} & {\textbf{77.83$\pm$0.62}} \\
    \multicolumn{1}{r}{} & {\textit{Conv}} & {\textbf{53.32$\pm$0.79}} & {\textbf{72.05$\pm$0.69}} \\
    \midrule

    \multicolumn{4}{c} {{\textbf{Methods with Conv Embedding}}} \\
    \midrule
    {\textbf{Matching Nets}} & {\textit{Conv}} & {43.56$\pm$0.84} & {55.31$\pm$0.73} \\
    {\textbf{\tabincell{c}{Meta-Learn LSTM}}} & {\textit{Conv}} & {43.44$\pm$0.77} & {60.60$\pm$0.71} \\

    {\textbf{MAML}} & {\textit{Conv}} & {48.70$\pm$1.84} & {63.11$\pm$0.92} \\

    {\textbf{CovaMNet}} & {\textit{Conv}} & {51.19$\pm$0.76} & {67.65$\pm$0.63} \\
    {\textbf{SalNet}} & {\textit{Conv}} & {\textbf{57.45$\pm$0.88}} & {72.01$\pm$0.67} \\
    \bottomrule
    \end{tabular}%
    }
  \label{tab:addlabel}%
\end{table}%

\paragraph{Generalizing to other datasets}

To better reflect the generalization performance of the few-shot learning models, we evaluate the few-shot learning model on the completely different datasets. A new dataset that totally different from the training dataset may present data distribution shift \cite{Recht2019}, which will cause significant performance degradation of the model. According to section 3.1, the training classes and the testing classes do not share the same label space, but they still possess the same data distribution because of coming from the same dataset. In this section, we train the model on miniImageNet and conduct the testing on the novel datasets to evaluate the generalization capability. The experiment results show that our model outperforms previous work(Proto Net, Relation Net, and K-tuplet loss \cite{Li2019R}) on the three novel datasets, which demonstrates the superior generalization capacity of our approach.
\begin{table}[H]
  \newcommand{\tabincell}[2]{\begin{tabular}{@{}#1@{}}#2\end{tabular}}
  \centering
  \caption{The mean accuracies of the 5-way 1-shot and 5-shot accuracies (\%) on three fined-grained datasets using the model trained on miniImageNet, with 95\% confidence intervals. All the experiments are conducted with the same network for fair comparison.}
    \scalebox{0.82}{
    \begin{tabular}{p{0.8cm}p{0.6cm}p{1.5cm}p{1.5cm}p{1.5cm}p{1.5cm}}
    \toprule
    \multicolumn{2}{c}{\multirow{3}[2]{*}{{Dataset}}} & \multirow{3}[2]{*}{{Proto Net}} & \multirow{3}[2]{*}{{\tabincell{l}{Relation \\ Net}}} & \multirow{3}[2]{*}{{\tabincell{l}{K-tuplet \\ loss}}} &  \multirow{3}[2]{*}{{\textbf{ours}}} \\
    \multicolumn{1}{c}{} & \multicolumn{1}{r}{} & \multicolumn{1}{r}{} \\ \multicolumn{1}{r}{} & \multicolumn{1}{r}{} & \multicolumn{1}{r}{} \\

    \midrule
    \multicolumn{1}{l}{\multirow{2}[2]{*}{{\tabincell{l}{Stanford \\ Dog}}}} & {1shot} & 31.54$\pm$0.41 & 31.24$\pm$0.61 & 37.33$\pm$0.65 &  {\textbf{42.11$\pm$0.84}} \\
          & {5shot} & 47.84$\pm$0.48 & 42.47$\pm$0.68 & 49.97$\pm$0.66 &  \textbf{59.98$\pm$0.79} \\
    \midrule
    \multicolumn{1}{l}{\multirow{2}[2]{*}{{\tabincell{l}{Stanford  \\Car}}}} & {1shot} & {29.19$\pm$0.40} & 28.83$\pm$0.55 & 31.20$\pm$0.58 &  {\textbf{32.97$\pm$0.62}} \\
          & {5shot} & 38.00$\pm$0.42 & 35.43$\pm$0.58 & 47.10$\pm$0.62 &  \textbf{51.58$\pm$0.71} \\
    \midrule
    \multicolumn{1}{l}{\multirow{2}[2]{*}{{CUB200}}} & {1shot} & {37.55$\pm$0.51} & 38.30$\pm$0.71 & 40.16$\pm$0.68 &  {\textbf{45.11$\pm$0.78}} \\
          & {5shot} & 55.03$\pm$0.49 & 50.89$\pm$0.69 & 56.96$\pm$0.65 &  \textbf{64.14$\pm$0.71} \\
    \bottomrule
    \end{tabular}%
    }
\end{table}%
\begin{table*}[ht]
  \newcommand{\tabincell}[2]{\begin{tabular}{@{}#1@{}}#2\end{tabular}}
  \centering
  \caption{The mean accuracies of the 5-way 1-shot and 5-shot tasks on three fine-grained datasets, with 95\% confidence intervals.}
    \scalebox{0.92}{
    \begin{tabular}{lcccccc}
    \toprule
    \multirow{2}[7]{*}{{\textbf{Method}}} &
     \multicolumn{6}{c}{{}{}{}{\textbf{5-Way accuracy(\%)}}} \\
    \cmidrule{2-7}
     {} &  \multicolumn{2}{c}{{\textit{Stanford Dogs}}} & \multicolumn{2}{c}{{\textit{Stanford Cars}}} & \multicolumn{2}{c}{{\textit{CUB 200-2011}}} \\

    {} &  \multicolumn{1}{c}{{1-shot}} & \multicolumn{1}{c}{{5-shot}} & \multicolumn{1}{c}{{1-shot}} & \multicolumn{1}{c}{{5-shot}} & \multicolumn{1}{c}{{1-shot}} & \multicolumn{1}{c}{{5-shot}} \\
    \midrule
     \multicolumn{1}{l}{{\textbf{Matching Net}}} &

     \multicolumn{1}{c}{{35.80$\pm$0.99}} & \multicolumn{1}{c}{{47.50$\pm$1.03}} & \multicolumn{1}{c}{{34.80$\pm$0.98}} & \multicolumn{1}{c}{{44.70$\pm$0.98}} & \multicolumn{1}{c}{{45.30$\pm$1.03}} & \multicolumn{1}{c}{{59.50$\pm$1.01}} \\
    \midrule
       \multicolumn{1}{l}{{\textbf{Proto Net}}} &

       \multicolumn{1}{c}{{37.59$\pm$1.00}} & \multicolumn{1}{c}{{48.19$\pm$1.03}} & \multicolumn{1}{c}{{40.90$\pm$1.01}} & \multicolumn{1}{c}{{52.93$\pm$1.03}} & \multicolumn{1}{c}{{37.36$\pm$1.00}} & \multicolumn{1}{c}{{45.28$\pm$1.03}} \\

    \midrule
      \multicolumn{1}{l}{{\textbf{Relation Net}}} &

      \multicolumn{1}{c}{{43.29$\pm$0.46}} & \multicolumn{1}{c}{{55.15$\pm$0.39}} & \multicolumn{1}{c}{{47.79$\pm$0.49}} & \multicolumn{1}{c}{{60.60$\pm$0.41}} & \multicolumn{1}{c}{{58.99$\pm$0.52}} & \multicolumn{1}{c}{{71.20$\pm$0.40}} \\

    \midrule
      \multicolumn{1}{l}{{\textbf{GNN}}} &

      \multicolumn{1}{c}{{46.98$\pm$0.98}} & \multicolumn{1}{c}{{62.27$\pm$0.95}} & \multicolumn{1}{c}{{55.85$\pm$0.97}} & \multicolumn{1}{c}{{71.25$\pm$0.89}} & \multicolumn{1}{c}{{51.83$\pm$0.98}} & \multicolumn{1}{c}{{63.69$\pm$0.94}} \\
    \midrule
      \multicolumn{1}{l}{{\textbf{CovaMNet}}} &

      \multicolumn{1}{c}{{49.10$\pm$0.76}} & \multicolumn{1}{c}{{63.04$\pm$0.65}} & \multicolumn{1}{c}{{56.65$\pm$0.86}} & \multicolumn{1}{c}{{71.33$\pm$0.62}} & \multicolumn{1}{c}{{52.42$\pm$0.76}} & \multicolumn{1}{c}{{63.76$\pm$0.64}} \\

    \midrule
      \multicolumn{1}{l}{{\textbf{LRPABN}}} &

      \multicolumn{1}{c}{{45.72$\pm$0.75}} & \multicolumn{1}{c}{{60.94$\pm$0.66}} & \multicolumn{1}{c}{{60.28$\pm$0.76}} & \multicolumn{1}{c}{{73.29$\pm$0.58}} & \multicolumn{1}{c}{{63.63$\pm$0.77}} & \multicolumn{1}{c}{{76.06$\pm$0.58}} \\

    \midrule
      \multicolumn{1}{l}{{\textbf{MattML}}} &

      \multicolumn{1}{c}{{54.84$\pm$0.53}} & \multicolumn{1}{c}{{71.34$\pm$0.38}} & \multicolumn{1}{c}{{{66.11$\pm$0.54}}} & \multicolumn{1}{c}{{{82.80$\pm$0.28}}} & \multicolumn{1}{c}{{\textbf{66.29$\pm$0.56}}} & \multicolumn{1}{c}{{80.34$\pm$0.30}} \\

    \midrule
      \multicolumn{1}{l}{{\textbf{DN4}}} &
      \multicolumn{1}{c}{{45.73$\pm$0.76}} & \multicolumn{1}{c}{{66.33$\pm$0.66}} & \multicolumn{1}{c}{{{61.51$\pm$0.85}}} & \multicolumn{1}{c}{{{89.60$\pm$0.44}}} & \multicolumn{1}{c}{{53.15$\pm$0.84}} & \multicolumn{1}{c}{{81.90$\pm$0.60}} \\

    \midrule
     \multicolumn{1}{l}{{{\textbf{Ours}}}} &

            \multicolumn{1}{c}{{\textbf{59.02$\pm$0.99}}} & \multicolumn{1}{c}{{\textbf{78.83$\pm$0.64}}} & \multicolumn{1}{c}{{\textbf{82.24$\pm$0.81}}} & \multicolumn{1}{c}{{\textbf{95.43$\pm$0.29}}} & \multicolumn{1}{c}{{{66.00$\pm$0.92}}} & \multicolumn{1}{c}{{\textbf{83.72$\pm$0.56}}} \\

    \bottomrule
    \end{tabular}%
  }
\end{table*}%

\paragraph{Few-shot Classification on fine-grained datasets}
Compared with the generic few-shot classification task, it’s more challenging to perform fine-grained few-shot (FGFS) classification due to the smaller inter-class and larger intra-class variations of the fine-grained datasets. However, since FGFS receives very little attention, most of the exiting few-shot learning methods do not report their performance on such fine-grained datasets. Therefore, we implement and evaluate our approach on fine-grained datasets. Meanwhile, we also report the DN4, CovaMNet, GNN \cite{Satorras2018}, Proto Net, MattML \cite{zhu2020multi}, LRPABN \cite{huang2020lowrank} to make a comparison. As table 3 shown, compared with those methods, our method achieves the best performance on three fine-grained datasets under both the 5-way 1-shot and 5-shot tasks. In more detail, on the Stanford Dogs dataset, our method gains 4.18\% and 15.79\% improvement respectively over the second place under 1-shot and 5-shot settings. On the Stanford Cars dataset, our method achieves the state of the art performance both on 1-shot and 5-shot, gaining 16.13\% and 5.83\% improvement over the second place. As for the CUB 200-2011 dataset, our method gains competitive accuracy under the 1-shot setting and best performance under the 5-shot setting. The result demonstrates that our methos helps to boost the fine-grained few-shot classification performance by utilizing the localization information.

\paragraph{Weakly-supervised object localization performance}
CUB200-2011 dataset is a benchmark for wsol task. It contains 200 categories of birds with 5994 training images and 5794 testing images. For each, it provides the bounding box for localization. We train our model on the training set without using any bounding box. During the meta-test phase, we predict the bounding box and the label for each input image. We use the Top-1 localization accuracy(Top-1 Loc) and localization accuracy with known ground truth class (GT-Known Loc). GT-Known is correct when the intersection over union (IoU) between the ground truth box and predicted box is 50\% or more. Top-1 Loc is correct when the Top-1 classification result (Top-1 Clas) and GT-Known Loc are both correct. We adopt the VGG-16 as our backbone for a fair comparison.

As table 4 shown, our method performs superior to the previous works both on Top-1 Loc acc and Top-1 Clas acc. Figure 5 also shows that the proposed method can locate greater regions of the objects than the CAM method.
\begin{table}[H]
\centering
\caption{Object localization preformance on CUB 200-2011}
\scalebox{0.9}{
\begin{tabular}{@{}ccccc@{}}
\toprule
\multirow{2}{*}{ Method} & \multirow{2}{*}{  Backbone} & \multicolumn{3}{c}{CUB-200-2011}                 \\ \cmidrule(l){3-5}
 &
   &
  \begin{tabular}[c]{@{}c@{}}Top-1\\  Loc (\%)\end{tabular} &
  \begin{tabular}[c]{@{}c@{}}Top-1\\  Clas (\%)\end{tabular} &
  \begin{tabular}[c]{@{}c@{}}GT-Known \\ Loc (\%)\end{tabular} \\ \cmidrule(l){1-5}
CAM                     & VGG-GAP                   & 34.41          & 67.55          & 57.96          \\
Acol                    & VGG-GAP                   & 45.92          & 71.9           & 59.3           \\
ADL                     & VGG-GAP                   & 52.36          & 65.27          & \textbf{75.41} \\
\textbf{Ours}           & VGG-GAP                   & \textbf{54.02} & \textbf{74.11} & 68.22          \\ \bottomrule
\end{tabular}
}
\end{table}

\begin{figure}
  \includegraphics[width=0.5\textwidth]{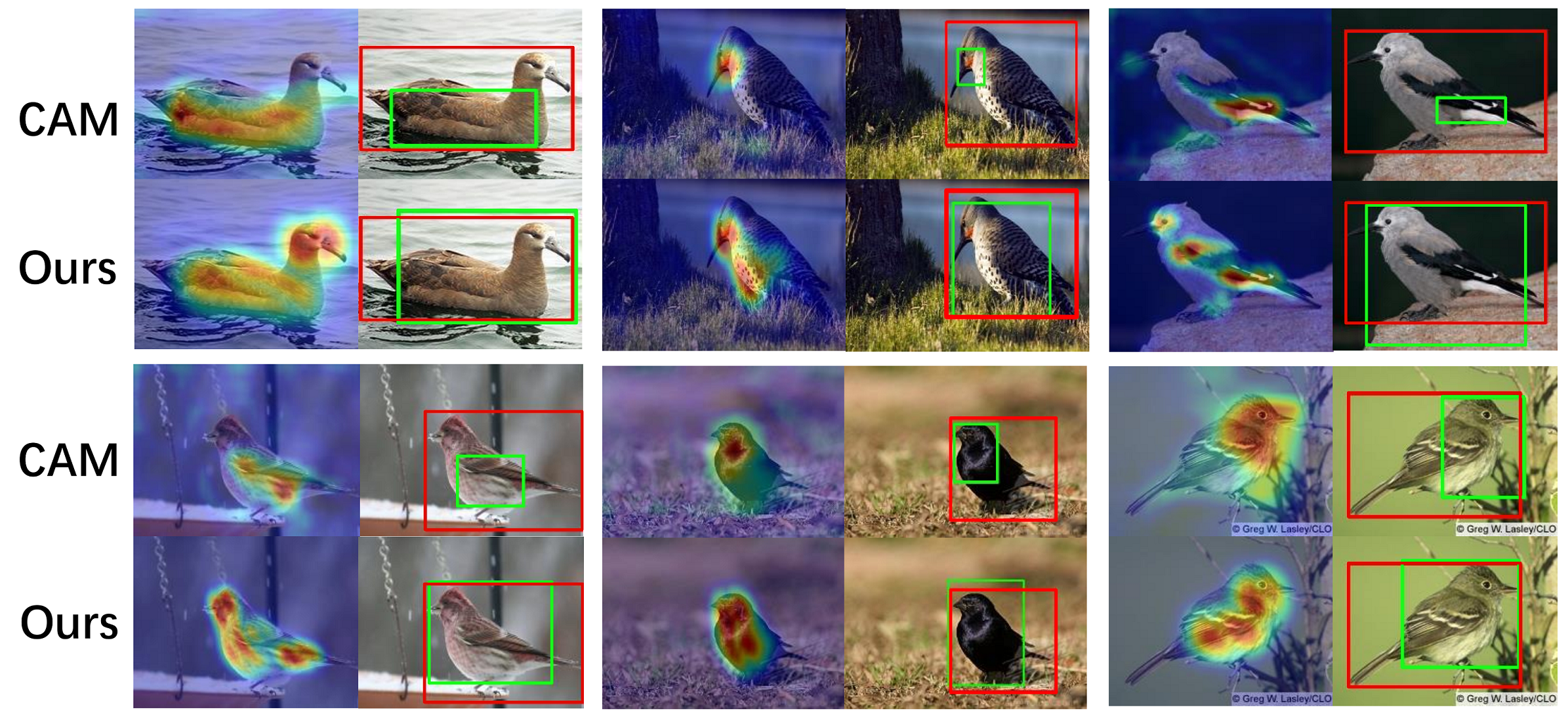}
  \caption{Compared with the CAM method on few-shot fine-grained datasets. Our method can locate more interval regions to improve localization performance. (ground-true bounding boxes are in red and the predicted are in green). Best vied in color. }
  \label{}
\end{figure}

\section{Discussion}
\subsection{Ablation study}
\paragraph{Influence of the backbone}
We execute the experiments to explore the influence of the backbone both on the few-shot learning dataset and fine-grained few-shot learning dataset. Conv-64 refers to a shallow network with 4 convolutional blocks and each block contains 64 filters with size $3 \times 3$, tailed with a max-pooling layer. The Resnet-12 refers to a ResNet-like network consisting of 4 residual blocks and each block contains 3 convolutional layers with $3 \times 3$ kernel. As the table shows, compares with Conv-64, ResNet-12 has made a significant improvement in all data sets under both 1-shot and 5-shot settings.
\begin{table}

  \newcommand{\tabincell}[2]{\begin{tabular}{@{}#1@{}}#2\end{tabular}}
  \centering
  \caption{Comparision the performance of different backbone (Conv-64 and ResNet-12) in our method. The mean accuracies of the 5-way 1-shot and 5-shot accuracies (\%) on three fined-grained datasets and miniImageNet dataset, with 95\% confidence intervals.}
   \scalebox{0.9}{
    \begin{tabular}{p{1cm}p{1cm}p{2cm}p{2cm}}
    \toprule
    \multicolumn{2}{c}{\multirow{2}[2]{*}{{Dataset}}} & \multirow{2}[2]{*}{{Conv-64}} & \multirow{2}[2]{*}{{\tabincell{l}{ResNet-12}}}  \\
     \multicolumn{1}{r}{} & \multicolumn{1}{r}{} & \multicolumn{1}{r}{} & \multicolumn{1}{r}{} \\

    \midrule
    \multicolumn{1}{l}{\multirow{2}[2]{*}{{\tabincell{l}{Stanford  Dog}}}} & {1-shot} & 50.77$\pm$0.91 & {\textbf{59.02$\pm$0.99}}\\
          & {5-shot} & 72.11$\pm$0.72 & {\textbf{78.83$\pm$0.64}}\\
    \midrule
    \multicolumn{1}{l}{\multirow{2}[2]{*}{{\tabincell{l}{Stanford   Car}}}} & {1-shot} & {58.58$\pm$0.85} & {\textbf{82.24$\pm$0.81}}\\
          & {5-shot} & 89.57$\pm$0.43 & {\textbf{95.43$\pm$0.29}} \\
    \midrule
    \multicolumn{1}{l}{\multirow{2}[2]{*}{{CUB 200-2011}}} & {1-shot} & {60.52$\pm$0.90} & {\textbf{66.00$\pm$0.92}} \\
          & {5-shot} & 80.36$\pm$0.61 & {\textbf{83.72$\pm$0.56}} \\

    \midrule
    \multicolumn{1}{l}{\multirow{2}[2]{*}{{mini-ImageNet}}} & {1-shot} & {53.32$\pm$0.79} & {\textbf{58.11$\pm$0.86}} \\
          & {5-shot} & 72.55$\pm$0.86 & {\textbf{77.83$\pm$0.62}} \\
    \bottomrule
    \end{tabular}%
    }
  \label{tab:addlabel}%
\end{table}%
\paragraph{Influence of the SAC and SAM}
To demonstrate the influence of various modules, we perform the ablation study and the results are shown in Table 4. Firstly, we replace the SAM with  Euclidean distance (ED) both in the w/SAC and w/o SAC combination. The implementation ED classifier is similar to the prototype network, where we use the vector flatted from the embedded feature to represent the embedded feature. The (w/SAC+SAM) outperforms (w/SAC+ED) in different settings, especially with the $Resnet$ backbone, gaining approximately 25.11\% improvement in 5-shot and 14.37\% in 1-shot, which demonstrates that the proposed semantic alignment distance can work well on few-shot classification. Secondly, without the SAC module to select the useful deep descriptor, the accuracy drops down according to the (w/SAC+SAM) and (w/o SAC+SAM). However, according to the (w/SAC+ED) and (w/o SAC+ED), it shows that the SAC can not work well with ED. This may because the ED can not utilize the semantic information provided by SAC. The ablation study shows our scheme can accurately utilize the weakly object localization to improve the few-shot learning performance.


\begin{table}
  \newcommand{\tabincell}[2]{\begin{tabular}{@{}#1@{}}#2\end{tabular}}
  \centering
  \caption{Testing ecah module of the proposed method during training on miniImageNet. The mean accuracies of the Resnet and Conv embedding module, with 95\% confidence intervals.}
  \scalebox{0.85}{
    \begin{tabular}{lccc}
    \toprule
    \multirow{2}[4]{*}{{\textbf{\tabincell{l}{Module \\Combination}}}} &
    \multirow{2}[4]{*}{{\textbf{Embedding}}} & \multicolumn{2}{c}{{\textbf{Acc \%}}} \\
    \cmidrule{3-4}
    \multicolumn{1}{l}{{}} & \multicolumn{1}{r}{{}} & {1-shot} & {5-shot} \\
    \midrule
    \multicolumn{1}{l}{\multirow{2}[2]{*}{{\textbf{\tabincell{l}{w/SAC+SAM}}}}} & {\textit{Resnet}} & {\textbf{58.11$\pm$0.86}} & {\textbf{77.83$\pm$0.62}} \\
          & {\textit{Conv}} & {\textbf{53.32$\pm$0.36}} & {\textbf{72.05$\pm$0.69}} \\
    \midrule
    \multicolumn{1}{l}{\multirow{2}[2]{*}{{\textbf{\tabincell{l}{w/SAC+ED}}}}} & {\textit{Resnet}} & {43.74$\pm$0.63} & {52.72$\pm$0.75} \\
          & {\textit{Conv}} & {41.45$\pm$0.67} & {53.32$\pm$0.72} \\
    \midrule
    \multicolumn{1}{l}{\multirow{2}[2]{*}{{\textbf{\tabincell{l}{w/o SAC+SAM}}}}} & {\textit{Resnet}} & {54.27$\pm$0.42} & {73.64$\pm$0.36} \\
          & {\textit{Conv}} & {51.24$\pm$0.74} & {69.72$\pm$0.64} \\
    \midrule
    \multicolumn{1}{l}{\multirow{2}[2]{*}{{\textbf{\tabincell{l}{w/o SAC+ED}}}}} & {\textit{Resnet}} & {51.15$\pm$0.85} & {69.02$\pm$0.75} \\
          & {\textit{Conv}} & {49.42$\pm$0.78} & {68.20$\pm$0.66} \\
    \bottomrule
    \end{tabular}%
    }
  \label{tab:addlabel}%
\end{table}%

\subsection{Visualization}
In this section, we visualize the class activation map of the input from the few-shot dataset and fine-grained few-shot datasets. Our method adopts the VGG-16 as the backbone to generate the class activation map for the input images. More specifically, we use the proposed SAC module to replace the last pooling layer and three fully connected layers of VGG16. Noted that the SAC module is a fully convolutional architecture, which means it can handle any size of the input easily. The model is trained end-to-end only on the training set. In the meta-test phase, we produce the cam for each novel images. As Figure 4 shown, compared with CAM, our model can more integral localize the key object both in the miniImageNet and the fine-grained datasets. It is worth mentioning that the model can generalize to novel categories (especially on fine-grained datasets) well since it can produce the cam for novel classes very well. This may because the unseen class always contains similar regions to the training set and the classifier will classify the novel sample to the most similar class in the training set.
\begin{figure}
  \centering
  \includegraphics[width=0.5\textwidth]{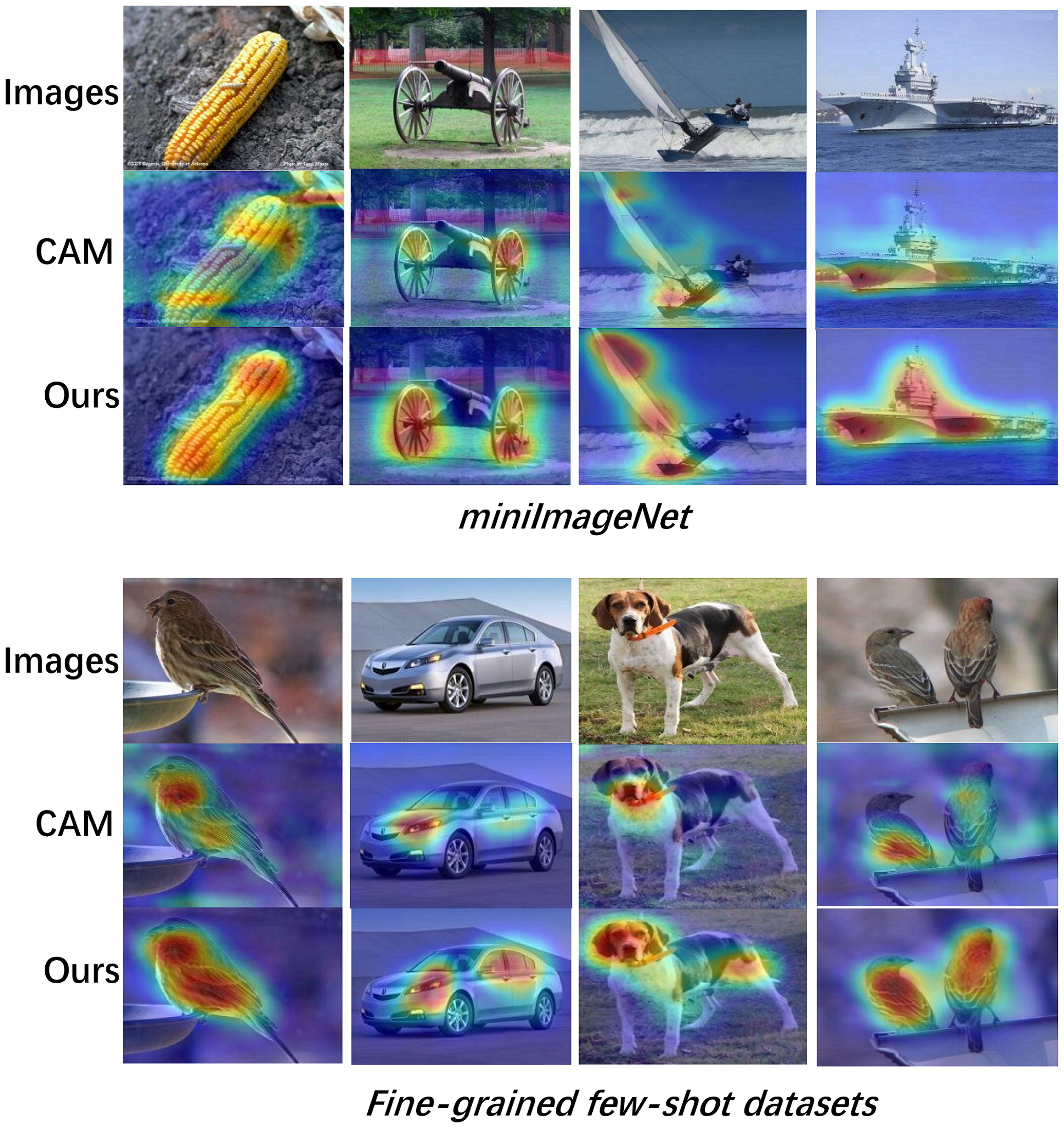}
  \caption{The class activation map generated by VGG-GAP-CAM and our method on miniImageNet dataset and three fine-grained few-shot datasets. The first row is input images, the second row is the class activation map generated by the CAM method, and the third row is the class attention map generate by ours. Compared with CAM, our method can capture more interval regions of the objects both in the miniImageNet and fine-grained datasets.}
  \label{}
\end{figure}

\section{Conclusions}
This paper proposes a method that can deal with both the few-shot classification and fined-grained few-shot classification well. The proposed method introduces the SAC module to localize the key objects, and more importantly selecting the useful deep descriptors for classification and fine-grained classification. The SAM module can align the semantic content between the query images and the support images by performing the NN algorithm over each selected deep descriptor. Extensive experiments show the proposed method obtains superior performance over state-of-the-art methods on both few-shot classification and fine-grained few-shot classification tasks. Furtherly, the ablation study shows that only our scheme can accurately utilize the subtle and local information to boost the performance of classification. The visualization shows the SAC module can localize the interval objects, which explains the high accuracies of our method.

\newpage

\bibliographystyle{named}

\bibliography{ijcai20}

\end{document}